\documentclass[runningheads]{llncs}
\usepackage[T1]{fontenc}
\usepackage{graphicx,verbatim}
\usepackage{amsmath, amssymb}
\usepackage{bbding}
\usepackage{pifont}

\begin{document}
%
\title{Cross-Modality Masked Learning for Survival Prediction in ICI Treated NSCLC Patients}
%

\author{Qilong Xing\inst{1} \and Zikai Song\inst{1} \thanks{Corresponding author: skyesong@hust.edu.cn} \and Bingxin Gong\inst{2} \and Lian Yang\inst{2} \and \\ Junqing Yu\inst{1} \and Wei Yang\inst{1}}
\authorrunning{Q. Xing et al.}
%
\institute{School of Computer Science and Technology \and Department of Radiology, Union Hospital, Tongji Medical College, \\Huazhong University of Science and Technology, Wuhan, China\\
\email{\{qlxing, skyesong\}@hust.edu.cn}}

\maketitle              
\begin{abstract}
Accurate prognosis of non-small cell lung cancer (NSCLC) patients undergoing immunotherapy is essential for personalized treatment planning, enabling informed patient decisions, and improving both treatment outcomes and quality of life. However, the lack of large, relevant datasets and effective multi-modal feature fusion strategies pose significant challenges in this domain. 
To address these challenges, we present a large-scale dataset and introduce a novel framework for multi-modal feature fusion aimed at enhancing the accuracy of survival prediction. The dataset comprises 3D CT images and corresponding clinical records from NSCLC patients treated with immune checkpoint inhibitors (ICI), along with progression-free survival (PFS) and overall survival (OS) data.
We further propose a cross-modality masked learning approach for medical feature fusion, 
consisting of two distinct branches, each tailored to its respective modality: a Slice-Depth Transformer for extracting 3D features from CT images and a graph-based Transformer for learning node features and relationships among clinical variables in tabular data. The fusion process is guided by a masked modality learning strategy, wherein the model utilizes the intact modality to reconstruct missing components. This mechanism improves the integration of modality-specific features, fostering more effective inter-modality relationships and feature interactions.
Our approach demonstrates superior performance in multi-modal integration for NSCLC survival prediction, surpassing existing methods and setting a new benchmark for prognostic models in this context.


\end{abstract}
\section{Introduction}
Lung cancer remains one of the leading causes of death worldwide, with non-small cell lung cancer (NSCLC) accounting for approximately 85\% \cite{85_per}. 
Immunotherapy, represented by immune checkpoint inhibitors (ICIs), targets the anti-programmed cell death protein 1 (PD-1), anti-programmed death-ligand 1 (PD-L1), and anti-cytotoxic T-lymphocyte-associated protein 4 (CTLA-4) pathways to enhance the body’s anti-tumor immune response, significantly improving the treatment of advanced or metastatic NSCLC \cite{ici}. 
However, many patients do not benefit from ICIs \cite{no_response}. It has been reported that only around 20\% of patients respond to ICIs, while an excessive immune response can disrupt immune tolerance, resulting in immune-related adverse events (irAEs) \cite{iraes}. 
Accurate prognosis in immunotherapy is crucial, as it enables healthcare providers to predict disease progression and tailor treatment plans accordingly. However, the lack of comprehensive datasets poses a significant challenge.

Regarding techniques used in survival prediction, recent advancements have shifted from traditional methods, such as the Kaplan-Meier estimator \cite{kaplan} and Cox proportional hazards regression \cite{cox}, to neural network-based approaches \cite{qu2024multi,kim2024llm,cai2024survival}. While single-modality methods \cite{survnet,haa} have shown promise, recent research \cite{cox_pasnet,page_net,mome,jiang2024multimodal,saeed2024survrnc} increasingly adopts multimodal strategies to leverage complementary information from diverse data sources. 
Compared to omics data, clinical data and medical images are generally more accessible. DAFT \cite{daft} introduces a dynamic affine module to fuse clinical data with 3D image features, while DOF \cite{dof} proposes an orthogonal loss function to encourage distinct modality representations. Additionally, Multisurv \cite{multisurv} combines clinical and whole-slide image features using an attention-based summation approach. However, these methods often rely on simplistic feature fusion techniques, which do not fully capture the complex inter-relationships between modalities, thus limiting the potential synergy of multimodal features.

To address these challenges, we propose two main contributions. First, we curate a large dataset comprising 3D CT images and corresponding clinical records of immunotherapy patients. This dataset includes progression-free survival (PFS) and overall survival (OS) data, which are essential for developing a robust survival prediction model for immunotherapy prognosis. Second, we introduce a cross-modality masked learning approach to enhance feature fusion, inspired by recent advancements in masked learning methods \cite{mae}. Specifically, we use a 3D visual transformer to encode the 3D CT images and employ a graph transformer to extract clinical features from the tabular data. The feature interaction and fusion are further strengthened by the cross-modality masked learning approach, which requires the model to complete missing information from one modality using the intact features from the other modality. 
Results confirm our method's effectiveness in multi-modal fusion and survival analysis.

\section{Method}
\begin{figure*}[t]
  \centering
    \includegraphics[width=1\linewidth]{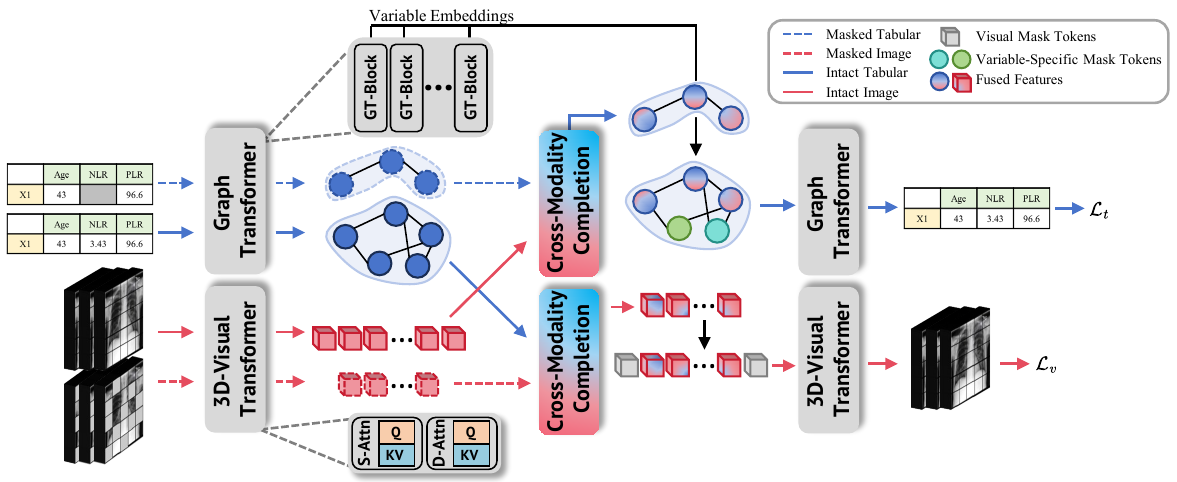}
    \caption{
During pretraining, both intact and masked versions of each modality are input into their respective branches. In the multi-modal completion process, the masked modality integrates features from the intact version of the other modality, which are then passed into the decoder for reconstruction. In the visual decoder, a learnable mask token is used, while in the graph transformer decoder, clinical variable-specific features from the encoder serve as masked tokens.
    }
    \label{fig:pipeline}
  \hfill
\end{figure*}

The proposed framework consists of two distinct branches: one dedicated to processing tabular clinical data and the other focused on 3D CT images. 
Our method follows a two-stage training procedure. The first stage involves pretraining the modality-specific encoders, while the second stage fine-tunes additional multi-layer perceptron (MLP) module for the survival prediction task. The overall architecture of the pretraining process is illustrated in Fig.\ref{fig:pipeline}.

\subsection{Visual Branch}
We employ a 3D visual transformer to extract image features from CT scans, drawing inspiration from recent advancements in video transformer models \cite{phenaki}. 
Given an input 3D CT image $\mathcal{I} \in \mathbb{R}^{H \times W \times D}$ and predefined patch sizes $p_1 \times p_2 \times p_3$, the model first reshapes the image into non-overlapping patches and transform to patch embeddings $ \mathcal{V} \in \mathbb{R}^{hw \times d \times C}$, where $h=\lfloor \frac{H}{p_1} \rfloor$, $w=\lfloor \frac{W}{p_2} \rfloor$, $d=\lfloor \frac{D}{p_3} \rfloor$, and $C=p_1 \times p_2 \times p_3$. 
The transformer encoder consists of a slice-based transformer (ST) block and a depth-based (DT) transformer block. 
In ST block, the slice-based attention (S-Attn) operates on the transposed patch embeddings $\mathcal{V}^T \in \mathbb{R}^{d \times hw \times C}$, using self-attention to capture global context across patches along the second dimension. The processed features are then transposed back and passed to the DT blocks, where the depth-based attention (D-Attn) facilitates information interaction along the depth dimension. The feature is then flattened for further computation:
\begin{equation}
    \mathcal{V}^{'} = \mathrm{D \text{-}Attn(S\text{-}Attn}(\mathcal{V}_p^T)^{T}) \in \mathbb{R}^{hwd \times c}, 
\end{equation}
During the masked learning procedure, image patches are randomly masked with a masking ratio $k$. Only the unmasked patches are fed into the model encoder for feature extraction. 
The decoder shares the same architecture as the encoder. 
To reconstruct the masked patches, following MAE \cite{mae}, the learnable masked embedding is combined with the visible patches and fed into the decoder. The reconstruction loss $\mathcal{L}_v$ is computed as the mean squared error (MSE) between the predicted and original patches.

\subsection{Tabular Branch}
We adapt the graph-based transformer T2G \cite{t2g} for our masked learning procedure. Categorical variables are encoded using learnable embeddings, while numerical variables are mapped via a linear module. All features are combined as input to the graph transformer. 
We treat each clinical variable as a node in the graph, and construct the relationships between variables as edges. Following the T2G framework, the graph edges built in each Graph Transformer block (GT-Block) consist of two components: one encodes the relationships based on specific sample features using a learnable relation diagonal matrix $R \in \mathbb{R}^{d\times d}$, and the other encodes the global relationships between variables using learnable embeddings $\mathcal{Z}=\{z_1,...,z_d\} \in \mathbb{R} ^{d \times c}$ for each clinical variable, where $d$ denotes the number of clinical variables and $c$ represents the feature channel size. The final graph is formed by combining these two types of edges.:
\begin{equation}
    G_s=\mathcal{X} R \mathcal{X}^T ,
    \quad
    G_z=\mathcal{Z} \mathcal{Z}^T,
    \quad
    G = \mathrm{Softmax}(G_s+G_z) \in \mathbb{R}^{d \times d},
\end{equation}
where $\mathcal{X}=\{x_1,...,x_n\} \in \mathbb{R}^{d \times c}$ represents the node features. 
The graph is used to update the node features, resulting in the processed features $\mathcal{X}^{'}$ obtained by the encoder.

In the masked learning procedure, $\lfloor dk \rfloor$ clinical variables are randomly masked per sample. Only the unmasked variables $\mathcal{X}_{[U]}$ are processed by the encoder, and edges associated with the masked nodes are removed, ensuring interaction occurs only between the remaining nodes.

Unlike visual features, graph features are less influenced by positional information. 
Therefore, simply combining position embeddings with a learnable masked embedding, as done in MAE \cite{mae}, is inadequate for providing the necessary information to facilitate effective reconstruction of masked clinical variables.
To address this, we propose collecting the embeddings for each clinical variable, $\mathcal{Z}$, from all $n$ blocks in the encoder and fusing them using a learnable matrix $W_z \in \mathbb{R}^{nc \times c}$ to construct a specific masked embedding for each clinical variable:
\begin{equation}
\begin{aligned}
    \mathcal{Z}_{a} = \mathrm{Concat}([\mathcal{Z}_1, ..., \mathcal{Z}_n]) \in \mathbb{R}^{d \times nc},
    \quad
    \mathcal{X}_{[M]} = \mathcal{Z}_{a}W_z \in \mathbb{R}^{d \times c},
\end{aligned}
\end{equation}
During reconstruction, the embedding for each masked variable is retrieved and combined with unmasked features for processing in the decoder, which shares the same architecture as the encoder. 
For the tabular reconstruction loss $\mathcal{L}_x$, we propose using different loss functions for numerical and categorical variables. In the final output layer of the decoder, numerical features are mapped to a single value, and the MSE loss is applied. For categorical variables, the features are mapped to logits. If the variable has only two categories, binary cross-entropy (BCE) loss is used; otherwise, cross-entropy (CE) loss is employed.

\subsection{Cross-Modality Completion} 

To enhance multimodal fusion, we propose a cross-modality completion (CMC) process based on masked learning. The CMC process requires the masked modality to complete itself by extracting and combining features from the other modality, strengthening the relationship between them and improving the feature fusion process. The CMC process is illustrated in Fig.\ref{fig:pipeline}.

Specifically, we employ $m$ transformer layers to progressively integrate cross-modality features within each branch. For instance, given the tabular features after random variable masking, $\mathcal{X}_{[U]}^{'}$, and the intact visual features, the self-attention mechanism is first applied to the tabular data to enhance its features: 
\begin{equation}
\begin{aligned}
    Q_s=\mathcal{X}_{[U]}^{'(m-1)} &W_{s,q}^{x,m}, \quad
    K_s=\mathcal{X}_{[U]}^{'(m-1)} W_{s,k}^{x,m}, \quad
    V_s=\mathcal{X}_{[U]}^{'(m-1)} W_{s,v}^{x,m},
    \\
    &\mathcal{X}_{[S,U]}^{'(m)} = \mathrm{LN}( \mathrm{Self \text{-}Attn}(Q_s, K_s, V_s) + \mathcal{X}_{[U]}^{'(m-1)}),
\end{aligned}
\end{equation}
Here, $W_{s,q}^{x,m}$, $W_{s,k}^{x,m}$ and $W_{s,v}^{x,m}$ are learnable matrices in the self-attention mechanism at the $m$-th layer of the tabular branch, and $\mathrm{LN}$ denotes layer normalization. 
To explore and fuse visual features for the reconstruction of masked clinical variables, we use the tabular features as the Query, while the visual features act as the Key and Value in the cross-modality attention mechanism:
\begin{equation}
\begin{aligned}
    Q_c=&\mathcal{X}_{[S,U]}^{'(m)} W_{c,q}^{x,m}, \quad
    K_c=\mathcal{V}^{'} W_{c,k}^{x,m}, \quad
    V_c=\mathcal{V}^{'} W_{c,v}^{x,m},
    \\
    &\mathcal{X}_{[C,U]}^{''(m)} = \mathrm{LN}(\mathrm{Cross \text{-}Attn}(Q_c,K_c,V_c) + \mathcal{X}_{[S,U]}^{'(m)}),
\end{aligned}
\end{equation}
where $W_{c,q}^{x,m}$, $W_{c,k}^{x,m}$ and $W_{c,v}^{x,m}$ are learnable matrices in the cross-attention process. Subsequently, a Multi-Layer Perceptron (MLP) with a residual connection is applied to produce the final feature at layer $m$:
\begin{equation}
\begin{aligned}
    \mathcal{X}_{[U]}^{'(m)} = \mathrm{MLP}(\mathcal{X}_{[C,U]}^{''(m)})+\mathcal{X}_{[C,U]}^{''(m)},
\end{aligned}
\end{equation}
The output of the CMC is fed into the graph transformer decoder, where it is combined with the masked embeddings. The operation of the visual branch is symmetric and therefore omitted for simplicity.

\subsection{Survival Analysis}
To fine-tune for the survival analysis task, we remove the decoder parts of the two branches and keep the remaining modules frozen. The intact clinical data and 3D CT images are input for feature extraction. 
After feature fusion through the transformers in the CMC process, the visual feature $\mathcal{V}_{[C,U]}^{'(m)}$ is pooled and concatenated with the $[CLS]$ token of the tabular feature $\mathcal{X}_{[C,U]}^{'(m)}$. These combined features are then processed by a MLP module to generate the survival prediction. For fine-tuning, we adopt the widely used Cox loss \cite{deepsurv}.

\section{Experiments}

\begin{table}[t]
\footnotesize
\centering
\caption{Sample distribution of categorical variables. Sample numbers are provided in parentheses. 
Abbreviations: AD (Adenocarcinoma), SCC (Squamous Cell Carcinoma), SD (Stable Disease), PD (Progressive Disease), PR (Partial Response).
Abbreviations: Adenocarcinoma (AD), Squamous Cell Carcinoma (SCC), Stable Disease (SD), Progressive Disease (PD), Partial Response (PR).
}
\label{tab:cat_tab}
\begin{tabular}{cl}
\hline
\multicolumn{1}{c}{Categorical Variables} & \multicolumn{1}{c}{Distributions}  \\ \hline
Sex & Male (1796) vs Female (332) \\
Lung cancer stage & Stage 3 (780) vs Stage 4 (1332) \\
Pathological types & AD (897) vs SCC (1040) vs Others (190) \\
Diabetes &  With diabetes (208) vs Without diabetes (1920) \\ 
Hypertension & With Hypertension (527) vs Without Hypertension (1599) \\
Smoking & Smoking (1134) vs Not smoking (842) \\
Drinking & Drinking (512) vs Not drinking (1388) \\
Hyperlipidemia & With hyperlipidemia (437) vs Without Hyperlipidemia (1568) \\
Response Evaluation & SD (1037) vs PD (392) vs PR (699) \\
\hline
\end{tabular}
\end{table}

\subsection{Dataset}
\begin{figure*}[t]
  \centering
    \includegraphics[width=0.95\linewidth]{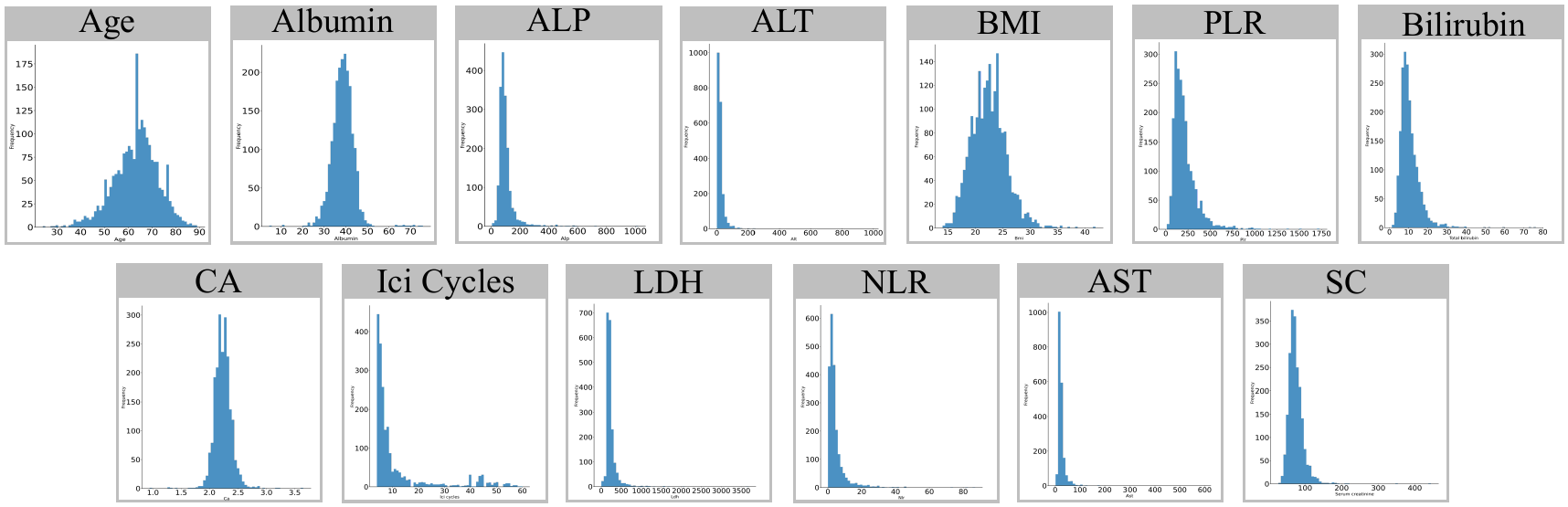}
    \caption{
    Sample distribution of numerical variables. Abbreviations: alkaline phosphatase (ALP), alanine aminotransferase (ALT), body mass index (BMI), platelet-to-lymphocyte ratio (PLR), calcium (CA), immune checkpoint inhibitor cycles (ICI cycles), lactate dehydrogenase (LDH), neutrophil-to-lymphocyte ratio (NLR), aspartate aminotransferase (AST), Serum creatinine (SC).
    }
    \label{fig:num_var_hist}
  \hfill
\end{figure*}

We curated a multimodal dataset for NSCLC immunotherapy patients, consisting of 3D CT images and clinical data from 2,128 samples. The CT images have a median resolution of \( 512 \times 512 \times 268 \) voxels and a median voxel spacing of \( 0.73 \times 0.73 \times 1.25 \) mm. The clinical data comprises 13 numerical (see Fig.~\ref{fig:num_var_hist}) and 9 categorical variables (see Table~\ref{tab:cat_tab}). The PFS and OS times are measured in months.

\subsection{Implementation Details}
We preprocess CT images by resampling them to a uniform voxel spacing of $0.75 \times 0.75 \times 1.5$ and center-cropping them to a size of $480 \times 480 \times 240$. For clinical data, missing numerical values are imputed with the mean, and categorical values with the most frequent class. We use five-fold cross-validation to evaluate PFS and OS predictions, with the Concordance Index (CI) as the metric \cite{autosurv}. 
The training consists of two stages. In pretraining, 
the default masking ratio is set to 50\%, the learning rate is $1e^{-4}$, and training lasts 80 epochs. In fine-tuning, the model is trained for 100 epochs with a learning rate of $1e^{-2}$. For comparison, we rerun competitive methods using their official implementations. All experiments are conducted on a single NVIDIA V100 GPU.

\begin{table}[t]
\footnotesize
\centering
\caption{The evaluation is based on two tasks: progression-free survival (PFS) prediction and overall survival (OS) prediction. * shows the statistical significance with significance level=0.05.}
\label{tab:main_result}
\begin{tabular}{lcc}
\hline
Methods           & PFS CI         & OS CI          \\ \hline
Cox \cite{cox}               & 0.653 $\pm$ 0.015         & 0.646  $\pm$ 0.016         \\
Interactive-Model \cite{duanmu} & 0.690 $\pm$ 0.018         & 0.687 $\pm$ 0.017         \\
FiLM \cite{film} & 0.681 $\pm$ 0.012          & 0.665 $\pm$ 0.020         \\
DAFT \cite{daft} & 0.686 $\pm$ 0.015         & 0.683  $\pm$ 0.016        \\ \hline
Ours & \textbf{0.701 $\pm$ 0.018}* & \textbf{0.705 $\pm$ 0.021}* \\ \hline
\end{tabular}
\end{table}

\subsection{Main Results}
We compare our method with the baseline Cox proportional hazards model \cite{cox}, which uses only tabular data, and other competitive methods that integrate both tabular and imaging data. The main results are shown in Table~\ref{tab:main_result}. 
Our method outperforms the baseline Cox model and other multimodal approaches on both PFS and OS prediction tasks, demonstrating the effectiveness of the CMC procedure. Notably, unlike methods that require retraining the entire model for each task (which takes hours), our method only fine-tunes a lightweight MLP module, taking less than 1 minute after pre-storing the features. This highlights the generalizability of our fused features across tasks.

\begin{table}[t]
\footnotesize
\centering
\caption{The ablation results.}
\label{tab:ablation}
\begin{tabular}{lcccc}
\hline
Methods & Image & Tabular & PFS CI         & OS CI          \\ \hline
VS   & \ding{52} & \ding{55} & 0.583 $\pm$ 0.011         & 0.591  $\pm$ 0.019         \\
VM   & \ding{52} & \ding{55} & 0.600 $\pm$ 0.015         & 0.604 $\pm$ 0.013         \\
TS & \ding{55} & \ding{52}  & 0.672 $\pm$ 0.010          & 0.643 $\pm$ 0.017         \\
TM &  \ding{55} & \ding{52}  & 0.670 $\pm$ 0.013          & 0.645 $\pm$ 0.014         \\
CS & \ding{52} & \ding{52}  & 0.685 $\pm$ 0.010         & 0.682  $\pm$ 0.019        \\ 
CM & \ding{52} & \ding{52}  & 0.689 $\pm$ 0.015         & 0.685  $\pm$ 0.010        \\ 
\hline
Ours w/o CVS & \ding{52} & \ding{52}  & 0.695 $\pm$ 0.017         & 0.696  $\pm$ 0.011        \\
Ours & \ding{52} & \ding{52} & \textbf{0.701 $\pm$ 0.018}* & \textbf{0.705 $\pm$ 0.021}* \\ \hline
\end{tabular}
\end{table}

\subsection{Model Analysis}
We conduct ablation studies to further evaluate the proposed method, with results shown in Table~\ref{tab:ablation}. First, we assess models based on individual modalities. Using only image data with the Cox loss (VS model) results in limited performance, while pretraining with the masked learning approach (VM model) improves performance. For tabular data, directly training with the Cox loss (TS model) yields similar performance to the masked learning-pretrained model, indicating that masked learning enhances the visual features more effectively than the tabular features.

For multimodal fusion, directly concatenating features from different branches and training with the Cox loss (CS model) outperforms single-modality models but performs worse than concatenating features enhanced by masked training for each branch separately (CM model). 
Our method achieves the highest accuracy. 
Additionally, we explore clinical-variable-specific (CVS) masked embeddings in the graph transformer, which prove more effective than using a single learnable mask token for different clinical variables.

\begin{figure*}[t]
  \centering
    \includegraphics[width=1\linewidth]{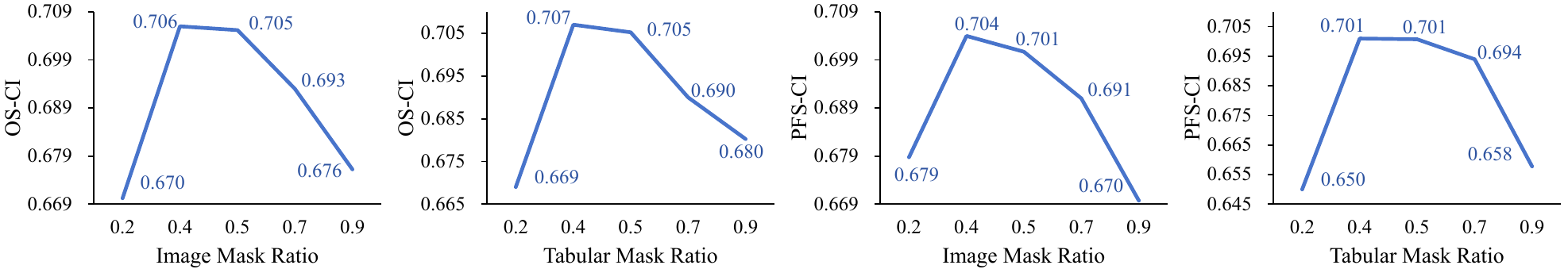}
    \caption{Effect of mask ratio on OS and PFS when varying the mask ratio for a single modality.}
    \label{fig:mask_ratio}
  \hfill
\end{figure*}

We also explore the effect of the mask ratio for different modalities, fixing one at 50\%. As shown in Fig.~\ref{fig:mask_ratio}, a moderate mask ratio for both modalities yields the best performance. Too low or too high a mask ratio for either modality prevents effective feature fusion, making the branch ineffective.

\begin{figure*}[t]
  \centering
    \includegraphics[width=1\linewidth]{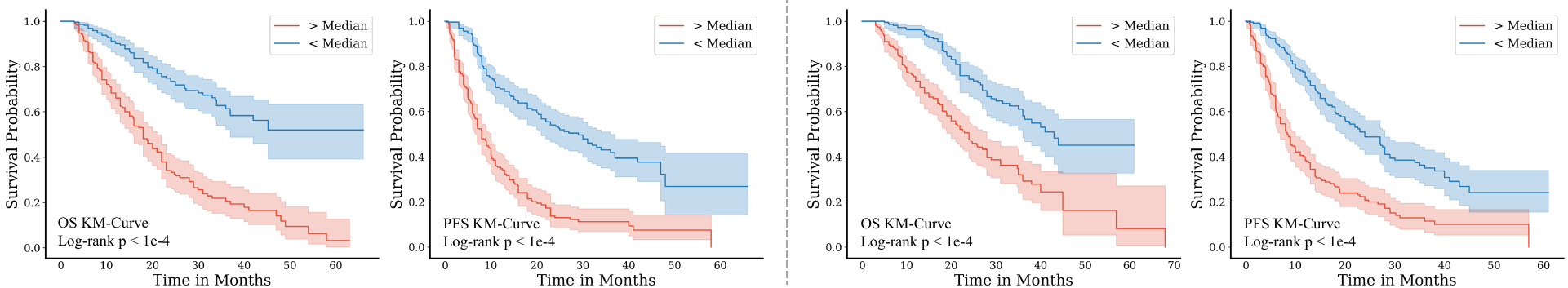}
    \caption{Kaplan-Meier curves and Log-rank tests of different CV folds.}
    \label{fig:km_curve}
  \hfill
\end{figure*}

In Fig.~\ref{fig:km_curve}, we present the Kaplan-Meier curve, splitting the test data using the median value of the predicted logits. The results show a very small p-value from the log-rank test, indicating that our method is effective for survival prediction.

\section{Conclusion}
Accurately predicting survival in ICI-treated NSCLC patients is essential for guiding treatment decisions. It allows for adaptive adjustments to treatment plans, ensuring personalized and effective care. 
To develop a robust model for survival prediction in NSCLC patients undergoing immunotherapy, we compile a large, relevant dataset and propose a cross-modality fusion method to optimize the use of multimodal data. Our experiments demonstrate that the proposed model effectively integrates features from multiple modalities and outperforms existing methods in survival prediction for NSCLC patients treated with immunotherapy.

\bibliographystyle{splncs04}
\bibliography{references}

\end{document}